\documentclass[runningheads]{llncs}
\usepackage{graphicx}
\usepackage{algorithm}
\usepackage{algorithmic}
\usepackage{amsmath}
\usepackage{array,tabularx}
\usepackage{amsmath,amssymb} 
\usepackage{color}
\usepackage[width=122mm,left=12mm,paperwidth=146mm,height=193mm,top=12mm,paperheight=217mm]{geometry}
\begin{document}

\title{YOLO3D: End-to-end real-time 3D Oriented Object Bounding Box Detection from LiDAR Point Cloud} 
\titlerunning{YOLO3D}

\authorrunning{ }

\author{Waleed Ali$^{1}$, Sherif Abdelkarim$^{1}$, Mohamed Zahran$^{1}$, Mahmoud Zidan$^{1}$, and Ahmad El Sallab$^{1}$
\thanks{$^{1}$Waleed Ali, Sherif Abdelkarim, Mahmoud Zidan, Mohamed Zahran, and Ahmad El-Sallab are with Valeo AI Research, Cairo, Egypt {\tt\small \{waleed.ali, sherif.abdelkarim, mahmoud.ismail-zidan.ext, mohamed.zahran, ahmad.el-sallab\}@valeo.com}}%
}
\institute{Valeo AI Research, Egypt}

\maketitle

\begin{abstract}
Object detection and classification in 3D is a key task in Automated Driving (AD). LiDAR sensors are employed to provide the 3D point cloud reconstruction of the surrounding environment, while the task of 3D object bounding box detection in real time remains a strong algorithmic challenge.
In this paper, we build on the success of the one-shot regression meta-architecture in the 2D perspective image space and extend it to generate oriented 3D object bounding boxes from LiDAR point cloud. Our main contribution is in extending the loss function of YOLO v2 to include the yaw angle, the 3D box center in Cartesian coordinates and the height of the box as a direct regression problem. This formulation enables real-time performance, which is essential for automated driving. Our results are showing promising figures on KITTI benchmark, achieving real-time performance (40 fps) on Titan X GPU.
\end{abstract}

\section{Introduction}

Automated Driving (AD) success is highly dependent on efficient environment perception. Sensors technology is an enabler to environment perception. LiDAR-based environment perception systems are essential components for homogeneous (same sensor type) or heterogeneous (different sensors types) fusion systems. The key feature of LiDAR is its physical ability to perceive depth at high accuracy. 

Among the most important tasks of the environment perception is Object Bounding Box (OBB) detection and classification, which may be done in the 2D (bird-view) or the 3D space. Unlike camera-based systems, LiDAR point clouds are lacking some features that exist in camera RGB perspective scenes, like colors. This makes the classification task from LiDAR only more complicated. On the other hand, depth is given as a natural measurement by LiDAR, which enables 3D OBB detections. The density of the LiDAR point cloud plays a vital role in the efficient classification of the object type, especially small objects like pedestrians and animals.

Real-time performance is essential in AD systems. While Deep Learning (DL) has a well-known success story in camera-based computer vision, such approaches suffer high latency in their inference path, due to the expensive convolution operations. In the context of object detection, rich literature exists that tackles the problem of real-time performance. Single shot detectors, like YOLO\cite{redmon2016you} and SSD \cite{liu2016ssd} are some of the best in this regard. 

In this paper, we extend YOLO V2\cite{redmon2017yolo9000} to perform 3D OBB detection and classification from 3D LiDAR point cloud (PCL). In the input phase, we feed the bird-view of the 3D PCL to the input convolution channels. The network architecture follows the meta-architecture of YOLO with architecture adaptation and tuning to match the nature of the sparse LiDAR input. The predictions include 8 regression outputs + classes (versus 5 regressors + classes in case of YOLO V2): the OBB center in 3D (x, y, z), the 3D dimensions (length, width and height), the orientation in the bird-view space, the confidence, and the object class label. Following the one-shot regression theme, we do not depend on any region proposal pipelines, instead, the whole system is trained end to end.

The main contributions of this work can be summarized as follows:

1-    Extending YOLO V2\cite{redmon2017yolo9000} to include orientation of the OBB as a direct regression task.

2-    Extending YOLO V2\cite{redmon2017yolo9000} to include the height and 3D OBB center coordinates (x,y,z) as a direct regression task.

3- Real-time performance evaluation and experimentation with Titan X GPU, on the challenging KITTI benchmark, with recommendations of the best grid-map resolution, and operating IoU threshold that balances speed and accuracy. 

The results evaluated on KITTI benchmark shows a clear advantage of the proposed approach, in terms of real-time efficiency (40 fps), and a promising accuracy. The rest of the paper is organized as follows: first, we discuss the related works, followed by a description of the proposed approach, and the combined one-shot loss for the 3D OBB, then we present, and discuss the experimental results on the KITTI benchmark dataset. Finally, we provide concluding remarks in section \ref{sec:conclusion}.

\begin{figure*}
 \centering
 \includegraphics[width=1\textwidth, keepaspectratio]{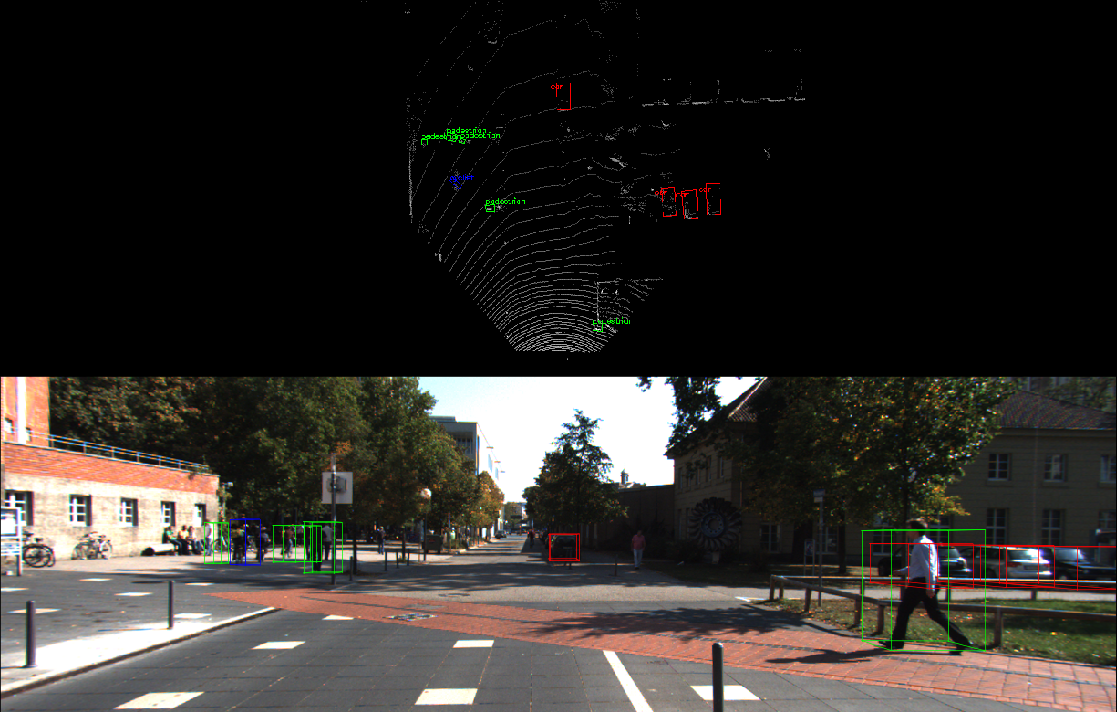}
  \caption{\label{fig: sample_output} Sample of the output shown in 3D and projected on the top view map}
\end{figure*}

\section{Related Work}
\label{sec:related}
\label{submission}
In this section, we summarize 3D object detection in autonomous driving for LiDAR point clouds. We then summarize related works in orientation prediction, which we use to predict the real angle of the vehicle. Finally, we discuss the implications of 3D object detection on the real-time performance.

\subsection{3D Object Detection}

There are three ways to do 3D object detection in terms of sensor type. Firstly, it is the LIDAR-only paradigm, which benefits from accurate depth information. Overall, these paradigms differ in data preprocessing. Some approaches project point cloud in 2D view (Bird-View, Front-View) such as\cite{li2016vehicle} and \cite{chen2017multi}; and some \cite{asvadi2017depthcn} convert the point cloud to a front view depth map. Others like \cite{li20173d} and \cite{zhou2017voxelnet}, convert the point cloud to voxels producing a discrete square grid.

The Second one is the camera-only paradigm; which works by adding prior knowledge about the objects' sizes, and trying to predict 3D bounding box using monocolor camera.\cite{mousavian20173d} and \cite{chabot2017deep} can produce highly accurate 3D bounding boxes using only camera images.\cite{chen20183d} uses stereo vision to produce high-quality 3D object detection. 

The LIDAR-camera fusion comes at the last. This paradigm tries to utilize the advantages of both paradigms mentioned above. The LIDAR produces accurate depth information, and the camera produces rich visual features; if we combine the output of the two sensors, we can have more accurate object detection and recognition. MV3D\cite{chen2017multi} fuses bird view, front view and the RGB camera to produce 3D vehicle detection. F-pointnet\cite{qi2017frustum} combines raw point cloud with RGB camera images. Object detection on RGB image produces a 2D bounding box which maps to a frustum in the point cloud. Then, 3D object detection is performed directly on frustum to produce accurate bounding boxes. However, fusing lidar data with camera suffers from adding more time complexity to the problem.

In this work, we are following the first paradigm of using only lidar point cloud projected as special bird view grid to keep the 3D information, more details will be discussed in section III-A.

\subsection{Orientation Prediction}
One approach in finding the orientation is introduced by MV3D\cite{chen2017multi}, where the orientation vector is assumed to be in the direction of the longer side of the box. This approach fails in regards to pedestrians because they don't obey this rule. 

Another approach is to convert the orientation vector to its component, as shown in \cite{ku2017joint} and \cite{simon2018complex}. AVOD\cite{ku2017joint} converts the orientation vector to sine and cosine.  Complex YOLO\cite{simon2018complex} converts the orientation vector to real and imaginary values. The problem with this is that the regression does not guarantee, or preserve any type of correlation between the two components of the angle.

\subsection{Real Time Performance}
Object detection is fundamental to automated driving, yet it suffers from computational complexity. There is a need to make the models as efficient as possible in terms of size and inference time maintaining a good accuracy.   

Some work has been done to tackle the efficiency of models, such as Squeeze-Net \cite{iandola2016squeezenet}, Mobile-Net \cite{howard2017mobilenets}, and Shuffle-Net \cite{zhang2017shufflenet}, and for object detection, there is Tiny YOLO and Tiny SSD \cite{wong2018tiny}. All these architectures are optimized for camera images, and they cannot easily be adapted to work on images produced from LiDAR point clouds. The reason is that, unlike camera images, LiDAR images consist of very sparse information. Vote3Deep \cite{engelcke2017vote3deep} performs 3D sparse convolution to take advantage of this sparsity.  

Extending YOLOv2 \cite{redmon2017yolo9000}, we include the orientation of the OBB as a direct regression task, unlike the work in \cite{simon2018complex}, which suggests two separate losses for the real and imaginary parts of the angle, without explicit nor implicit correlation between them, which may result in wrong or invalid angles in many cases.

In addition, in \cite{simon2018complex}, 3D OBB height and z-center are not a natural or exact output from the network, but rather a heuristic based on statistics and average sizes of the data. In this work, we extend YoLo v2\cite{redmon2017yolo9000} to include height and 3D OBB center as direct regression tasks. A sample of our output can be seen in Figure (\ref{fig: sample_output}), taken from KITTI benchmark test data.

\section{Approach}
\label{sec:approach}
\begin{figure*}
  \centering
  \includegraphics[width=0.9\textwidth, keepaspectratio]{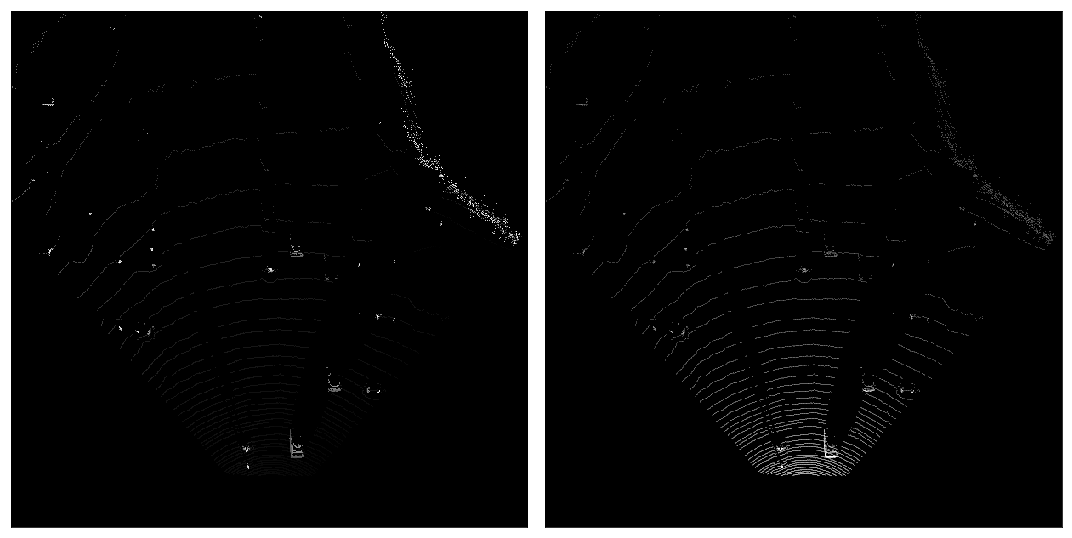}
  \caption{\label{fig: input_format} Sample of the input grid maps. Left: the height map. Right: the density map.}
\end{figure*}
\subsection{Point Cloud Representation}
 We project the point cloud to create a bird's eye view grid map. We create two grid maps from the projection of the point cloud as shown in Figure (\ref{fig: input_format}). The first feature map contains the maximum height, where each grid cell (pixel) value represents the height of the highest point associated with that cell. The second grid map represent the density of points. Which means, the more points are associated with a grid cell, the higher its value would be. The density is calculated using the following equation taken from MV3D paper \cite{chen2017multi}:
\begin{equation}\label{eqn: density}
min(1.0, \frac{log(N+1)}{log(64)})
\end{equation}
Where N is the number of points in each grid cell.

\subsection{Yaw Angle Regression}

The orientation of the bounding boxes has a range from -$\pi$ to $\pi$. We normalized that range to be from -1 to 1, and adapted our model to directly predict the orientation of the bounding box via a single regressed number. In the loss function, we compute the mean squared error between the ground truth and our predicted angle:
\begin{equation}\label{eqn: yaw}
\sum_{i=0}^{s^2}\sum_{j=0}^{B}L_{ij}^{obj} (\phi_i - \hat{\phi_i})^2
\end{equation}
  In our experimentation, we used using tanh as an activation for the angle output yaw (to bound the output between -1 and 1), but it did not offer an improvement over the linear activation.

\subsection{3D Bounding Box Regression}
\begin{figure}
    \centering
    \includegraphics[width=8cm, keepaspectratio]{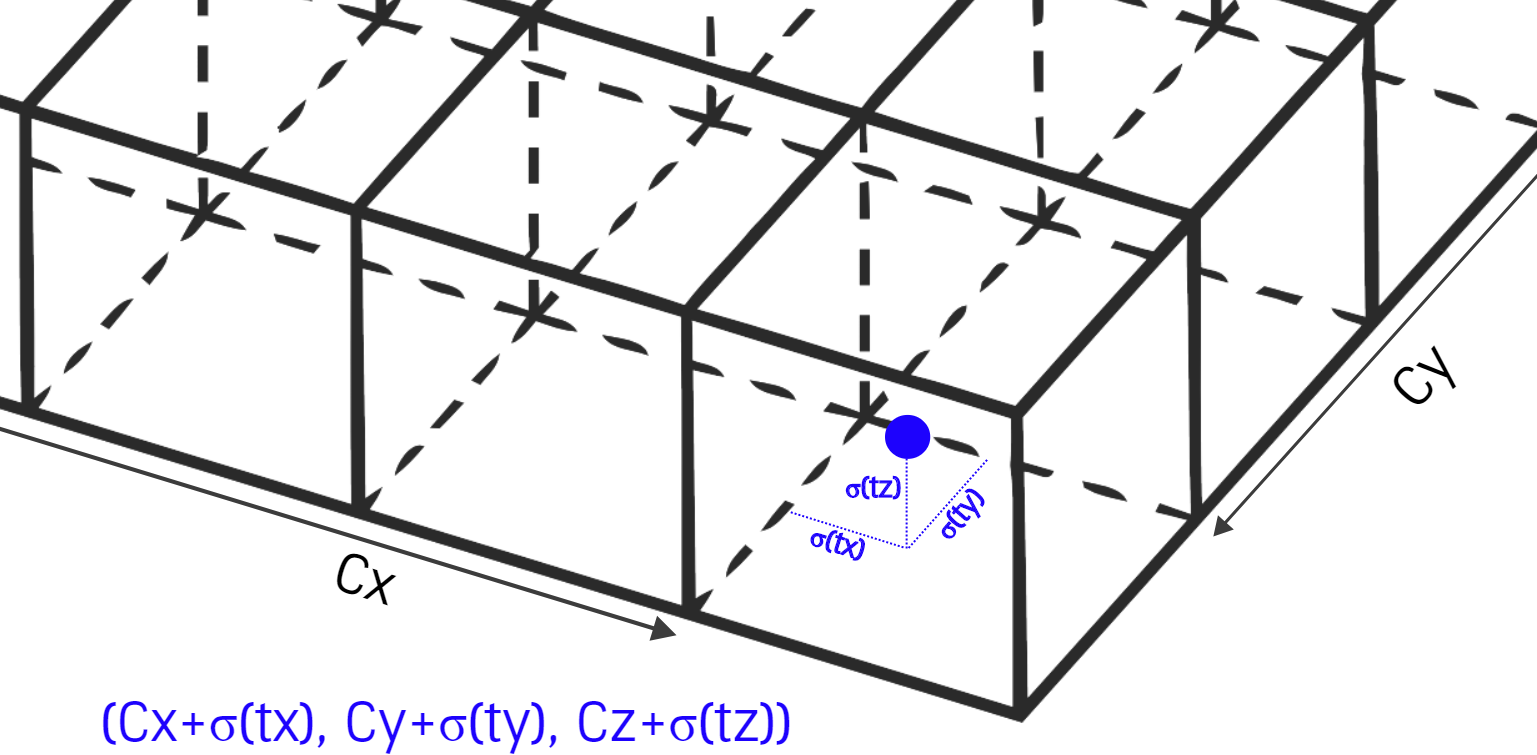}
    \caption{\label{fig: grid_3d} Sample of the grid output when extended to the third dimension where c\textsubscript{z} equals 0 since the grids are only one level high in the z dimension.}
  \end{figure}
We added two regression terms to the original YOLO v2\cite{redmon2017yolo9000} in order to produce 3D bounding boxes, the z coordinate of the center, and the height of the box.
The regression over the z coordinate in Eq. (\ref{3}) is done in a way similar to the regression of the x Eq. (\ref{1}) and y Eq. (\ref{2}) coordinates via a sigmoid activation function. 
\par While the x and y are regressed by predicting a value between 0 and 1 at each grid cell, locating where the point lies within that cell, the value of z is only mapped to lie within one vertical grid cell as illustrated in Figure (\ref{fig: grid_3d}). The reason for choosing to map z values to only one grid while x and y are mapped to several grid cells is that the variability of values in the z dimension are much smaller than that of the x and y (most objects have very similar box elevations).

The height of the box h Eq. (\ref{6}) is also predicted similarly to the width w in Eq. (\ref{4}) and length l in Eq. (\ref{5})

\begin{equation}\label{1}
	b_x=\sigma(t_x) + c_x
\end{equation}
\begin{equation}\label{2}
    b_y=\sigma(t_y) + c_y
\end{equation}
\begin{equation}\label{3}
    b_z=\sigma(t_z) + c_z
\end{equation}
\begin{equation}\label{4}
    b_w=p_we^{t_w}
\end{equation}
\begin{equation}\label{5}
    b_l=p_le^{t_l}
\end{equation}
\begin{equation}\label{6}
	b_h=p_he^{t_h}
\end{equation}
\subsection{Anchors Calculation}
In YOLO-v2 \cite{redmon2017yolo9000}, anchors are calculated using k-means clustering over the width and length of the ground truth boxes. The point behind using anchors, is to find priors for the boxes, onto which the model can predict modifications. The anchors must be able to cover the whole range of boxes that can appear in the data. In \cite{redmon2017yolo9000}, the model is trained on camera images, where there is a high variability of box sizes, even for the same object class. Therefore, calculating anchors using clustering is beneficial.
\par On the other hand, in the case of bird's eye view grid maps, there is no such high variability in box dimensions within the same object class (most cars have similar sizes). For this reason, we chose not to use clustering to calculate the anchors, and instead, calculate the mean 3D box dimensions for each object class, and use these average box dimensions as our anchors.

\subsection{Combined Loss for 3D OBB}
The loss for 3D oriented boxes is an extension to the original YOLO loss for 2D boxes. The loss for the yaw term is calculated as described in subsection B and Eq. (\ref{eqn: yaw}). The loss for the height is an extension to the loss over the width and length in (\ref{loss_equation}). Similarly, the loss for the z coordinate is an extension to the loss over the x and y coordinates, as shown in (\ref{loss_equation}). 
\par The total loss shown in Eq. (\ref{loss_equation}) is calculated as the scaled summation of the following terms: the mean squared error over the 3D coordinates and dimensions (x, y, z, w, l, h), the mean squared error over the angle, the confidence score, and the cross entropy loss over the object classes.
  
\begin{equation}\label{loss_equation}
\begin{split}
	&L=\lambda_{coor}\sum_{i=0}^{s^2}\sum_{j=0}^{B}L_{ij}^{obj} [(x_i-\hat{x_i})^2+(y_i-\hat{y_i})^2+(z_i-\hat{z_i})^2]\\
    &\qquad +\lambda_{coor}\sum_{i=0}^{s^2}\sum_{j=0}^{B}L_{ij}^{obj}[(\sqrt{w_i}-\sqrt{\hat{w_i}})^2 + (\sqrt{l_i}-\sqrt{\hat{l_i}})^2 \\
    &\qquad + (\sqrt{h_i}-\sqrt{\hat{h_i}})^2] \\
    &\qquad + \lambda_{yaw}\sum_{i=0}^{s^2}\sum_{j=0}^{B}L_{ij}^{obj}(\phi_i-\hat{\phi_i})^2 \\
    &\qquad + \lambda_{conf}\sum_{i=0}^{s^2}\sum_{j=0}^{B}L_{ij}^{obj}(C_i-\hat{C_i})^2 \\
    &\qquad + \lambda_{conf}\sum_{i=0}^{s^2}\sum_{j=0}^{B}L_{ij}^{noobj}(C_i-\hat{C_i})^2 \\
    &\qquad + \lambda_{classes}\sum_{i=0}^{s^2}\sum_{j=0}^{B}L_{ij}^{obj}\sum_{c\in classes}(p_i(c)-\hat{p}_i(c))^2
\end{split}
\end{equation}

Where:
$\lambda_{coor}:$ the weight assigned to the loss over the coordinates, $\lambda_{conf}:$ the weight assigned to the loss over predicting the confidence, $\lambda_{yaw}:$ the weight assigned to the loss over the orientation angle, $\lambda_{classes}:$ the weight assigned to the loss over the class probabilities, $L_{ij}^{obj}:$ a variable that takes the values of 0 and 1 based on whether there is a ground truth box in the ith and jth location. 1 if there's a box, and 0 otherwise, $L_{ij}^{noobj}:$ the opposite of the previous variable. takes the value of 0 if there's no object, and 1 otherwise, $x_i, y_i, z_i:$ the gound truth coordinates, $\hat{x_i}, \hat{y_i}, \hat{z_i}:$ the ground truth and predicted orientation angle, $\phi_i, \hat{\phi_i}:$ the ground truth and predicted orientation angle	  ..and so on, $C_i, \hat{C_i}:$ the ground truth and predicted confidence, $w_i, l_i, h_i:$ the ground truth width, height, and length of the box, $ \hat{w_i}, \hat{l_i}, \hat{h_i}:$ the predicted width, height, and length of the box and $p_i(c), \hat{p}_i(c:)$The ground truth and predicted class probabilities.

\section{Experiments and Results}
\label{sec:experiments}
\subsection{Network Architecture and Hyper Parameters}
Our model is based on YOLO-v2\cite{redmon2017yolo9000} architecture with some changes, as shown in Table \ref{table: network}.
\begin{enumerate}
\item We modified one max-pooling layer to change the down-sampling from 32 to 16 so we can have a larger grid at the end; this has a contribution in detecting small objects like pedestrians and cyclists.

\item We removed the skip connection from the model as we found it resulting in less accurate results.

\item We added terms in the loss function for yaw, z center coordinate, and height regressions to facilitate the 3D oriented bounding box detection.

\item Our input consists of 2 channels, one representing the maximum height, and the other one representing the density of points in the point cloud, computed as shown in Eq. (\ref{eqn: density}). 

\end{enumerate}
\begin{table}[ht!]
\centering
\caption{\label{table: network} Network Architecture}
\begin{tabular}{ |l|c|c|c| }\hline
{layer}&\makebox[3em]{filters}&\makebox[5em]{size}&\makebox[3em]{feature maps}\\\hline\hline
conv2d& 32 & (3 , 3) & 608x608x2 \\\hline 
maxpooling & - & (size 2 , stride 2) &  \\\hline 
conv2d & 64 & (3 , 3) &  \\\hline 
maxpooling & - & (size 2 , stride 2) &  \\\hline 
conv2d & 128& (3 , 3) & \\\hline 
conv2d & 64 & (3 , 3) &   \\\hline 
conv2d & 128& (3 , 3) &   \\\hline 
maxpooling & - & (size 2 , stride 1) &  \\\hline 
conv2d & 256& (3 , 3) &  \\\hline 
conv2d & 128& (3 , 3) &   \\\hline 
conv2d & 256 & (3 , 3) &  \\\hline 
maxpooling & - & (size 2 , stride 2) &  \\\hline 
conv2d & 512 & (3 , 3) &   \\\hline 
conv2d & 256 & (1 ,1) &   \\\hline 
conv2d & 512 & (3 , 3) &  \\\hline 
conv2d & 256 & (1 ,1) &   \\\hline 
conv2d & 512 & (3 , 3) &   \\\hline 
maxpooling & - & (size 2 , stride 2) &  \\\hline 
conv2d & 1024 & (3 , 3) &   \\\hline 
conv2d & 512 & (1 , 1) &   \\\hline 
conv2d & 1024 & (3 , 3) &   \\\hline 
conv2d & 512 & (1 , 1) &   \\\hline 
conv2d & 1024 & (3 , 3) &   \\\hline 
conv2d & 1024 & (3 , 3) &   \\\hline 
conv2d & 1024 & (3 , 3) &   \\\hline 
conv2d & 1024 & (3 , 3) &  \\\hline 
conv2d & 1024 & (1 ,1) & 38x38x33   \\\hline 
reshape & - & - &  38x38x3x11 \\\hline
\end{tabular}
\end{table}
\subsection{Dataset and Preprocessing}

We used KITTI benchmark dataset. The point cloud was projected in 2D space as a bird view grid map with a resolution of 0.1m per pixel, same resolution is used by MV3D\cite{chen2017multi}. 
\par The range represented from the LiDAR space by the grid map is 30.4 meters to right and 30.4 meters to the left, and 60.8 meters forward. Using this range with the above mentioned resolution of 0.1 results in an input shape of 608x608 per channel.
\par The height in the LiDAR space is clipped between +2m and -2m, and scaled to be from 0 to 255 to be represented as pixel values in the maximum height channel.
\par Since in KITTI benchmark only the objects that lies
on the image plane are labeled, we filter any points from the point cloud that lie outside the image plane. The rationale behind this, is to avoid giving the model contradictory information. Since objects lying on the image plane would need to be detected, while the ones lying outside that plane should be ignored, as they are not labeled. Therefore, we only include the points that lie within the image plane.

\begin{figure*}
\begin{minipage}{.33\textwidth}
  \centering
  \includegraphics[width=\linewidth, keepaspectratio]{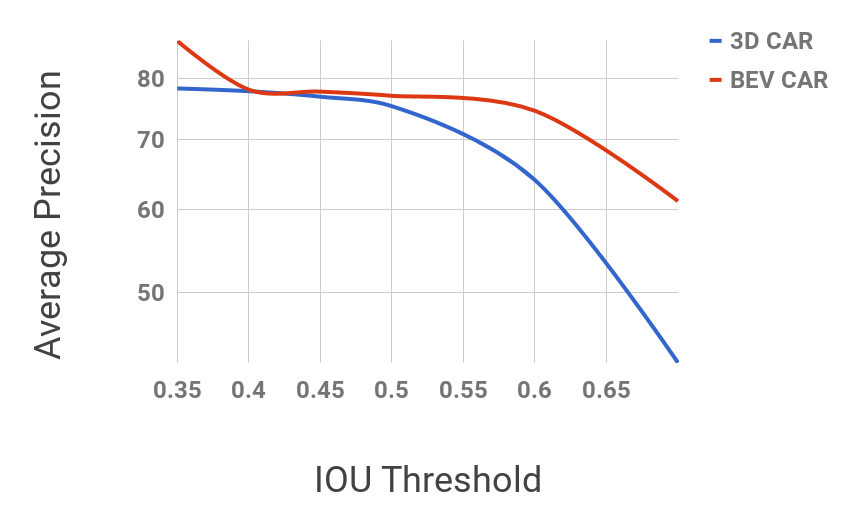}
  \caption{car}
  \label{fig:car}
\end{minipage}%
\begin{minipage}{.33\textwidth}
  \centering
  \includegraphics[width=\linewidth, keepaspectratio]{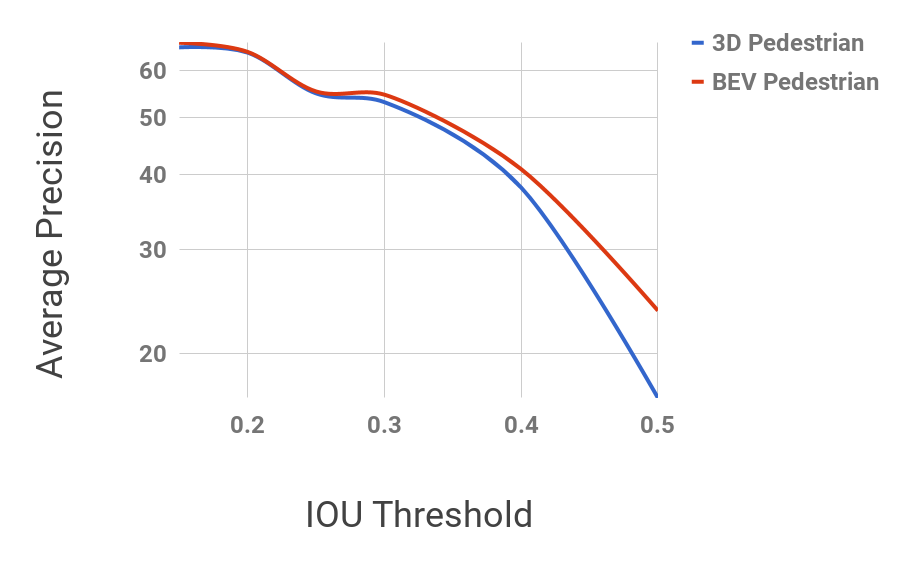}
  \caption{pedestrian}
  \label{fig:pedestrian}
\end{minipage}%
\begin{minipage}{.33\textwidth}
  \centering
  \includegraphics[width=\linewidth, keepaspectratio]{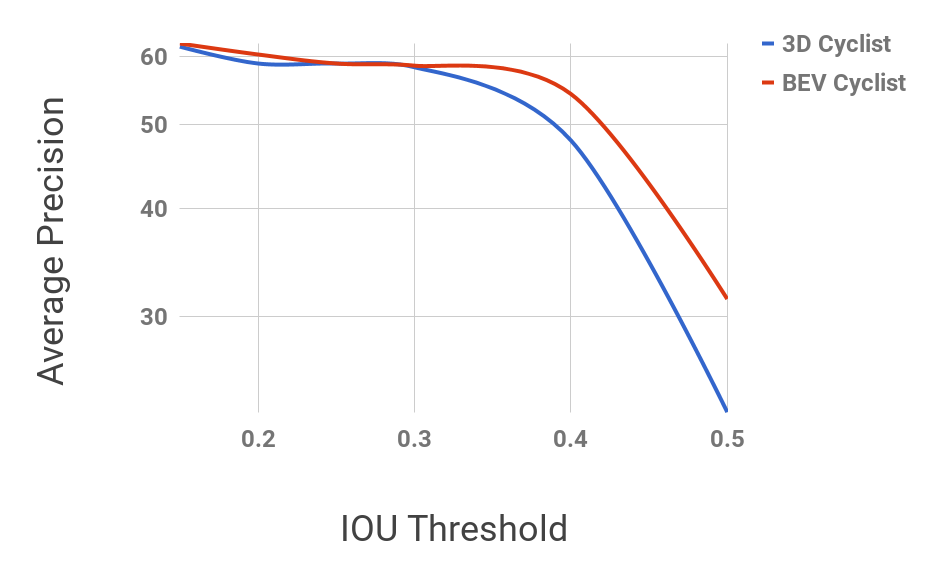}
  \caption{cyclist}
  \label{fig:cyclist}
\end{minipage}%
\caption{Performance against IOU threshold}
\label{fig: iou}
\end{figure*}
\subsection{Training}
The network is trained in an end-to-end fashion. We used stochastic gradient descent with a momentum of 0.9, and a weight decay of 0.0005. We trained the network for 150 epochs, with a batch size of 4.
\par Our learning rate schedule is as follows:
for the first few epochs, we slowly raise the learning rate from 0.00001 to 0.0001. If we start at a high learning rate, our model often diverges due to unstable gradients. We continue training with 0.0001
for 90 epochs, then 0.0005 for 30 epochs, and finally, 0.00005 for the last 20 epochs.

\subsection{KITTI Results and Error Analysis}

As discussed in \cite{farhadiyolov3}, and from the results reported in \cite{redmon2016you} and \cite{redmon2017yolo9000}, YOLO performs very well with the detection metric of mean average precision at IOU threshold of 0.5. This gives us an advantage over the previous work in 3D detection from point cloud in terms of speed with an accepted mAP, as shown in Figure (\ref{fig: iou}).

However, performance drops significantly as the IOU threshold increases indicating that we struggle to get the boxes perfectly aligned with the object, which is an inherited problem in all YOLO versions \cite{redmon2016you}, \cite{redmon2017yolo9000}, \cite{farhadiyolov3}. Fig \ref{fig: iou} shows that the model succeeds in detecting the objects but struggles with accurately localizing them.

Compared with the state of the art approaches on 3D object detection, such as MV3D\cite{chen2017multi}, which fails in detecting pedestrians and cyclists despite its relatively large, and complex multi view, multi sensor network, as well as, AVOD\cite{ku2017joint}, which dedicates a separate network for detecting cars, and one for pedestrians and cyclists, our proposed architecture can detect all objects from only a two channel bird view input, and with just one single network, achieving a real time performance of 40 fps , and a 75.3\% mAP on 0.5 IOU threshold for moderate cars. The precision and recall scores on our validation set (about 40\% of the KITTI training set) are shown in Table \ref{table: validation}.

\begin{figure}
  \centering
\includegraphics[width=0.8\linewidth, keepaspectratio]{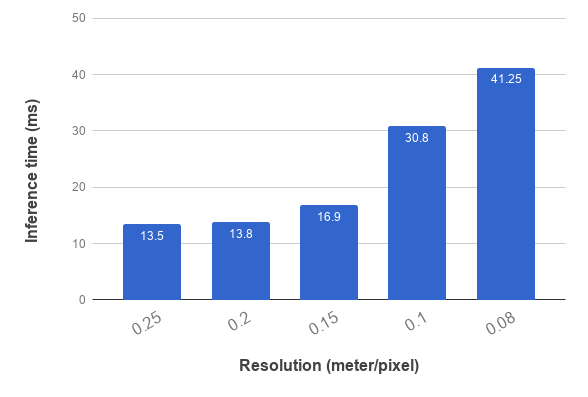}
  \caption{\label{fig: inference_time} Inference time at different resolutions}
\end{figure}

\begin{table}[ht!]
\centering
\caption{\label{table: validation} Validation Results}
\begin{tabular}{|l|l|l|}
\hline
label & precision & recall \\ 
\hline
pedestrian & 44.0\% & 39.2\% \\ 
\hline 
cyclist & 65.13\% & 51.1\% \\ 
\hline 
car & 94.07\% & 83.4\% \\ 
\hline 
\end{tabular}
\end{table}

\subsection{Effect of Grid Map Resolution}
\par Grid map resolution is a critical hyper-parameter that affect memory usage, time and performance. For instance, if we want to deploy the model on an embedded target, we have to focus on fast inference time with small input size, and reasonable performance.
\par The area of the grid map grows proportionally to the length or width of the grid map squared. This means increasing the resolution of the grid map, increases the area of the grid map (and thus the inference time) quadratically. This can be seen in Figure (\ref{fig: inference_time}), where there is a rapid increase in the inference time after the 0.15 meter/pixel mark, where only increasing the resolution by 0.05 meters/pixel (from 0.15 meters/pixel to 0.1 meter/pixel) causes the inference time to double from 16.9ms to 30.8ms.

\section{Conclusions}
\label{sec:conclusion}
In this paper we present real-time LiDAR based system for 3D OBB detection and classification, based on extending YOLO-v2\cite{redmon2017yolo9000}. The presented approach is trained end to end, without any pipelines of region proposals which ensure real time performance in the inference pass. The box orientation is ensured by direction regression on the yaw angle in bird-view. The 3D OBB center coordinates and dimensions are also formulated as a direct regression task, with no heuristics. The system is evaluated on the official KITTI benchmark at different IoU thresholds, with recommendation of the best operating point to get real time performance and best accuracy. In addition, the real time performance is evaluated at different grid-map resolutions. The results suggest that single shot detectors can be extended to predict 3D boxes while maintaining real-time performance; however this comes with a cost on the localization accuracy of the boxes.

\bibliographystyle{splncs}
\bibliography{eccv2018submission}
\end{document}